\pdfoutput=1

\RequirePackage{snapshot}
\documentclass[11pt]{article}

\usepackage{ACL2023}

\usepackage{times}
\usepackage{latexsym}

\usepackage[T1]{fontenc}

\usepackage[utf8]{inputenc}

\usepackage{microtype}

\usepackage{inconsolata}

\usepackage{amsmath,amsfonts,bm}

\def\eqref#1{equation~\ref{#1}}

\def\1{\bm{1}}

\DeclareMathAlphabet{\mathsfit}{\encodingdefault}{\sfdefault}{m}{sl}
\SetMathAlphabet{\mathsfit}{bold}{\encodingdefault}{\sfdefault}{bx}{n}

\DeclareMathOperator*{\argmax}{arg\,max}
\DeclareMathOperator*{\argmin}{arg\,min}

\usepackage{booktabs}
\usepackage{graphicx}
\usepackage{caption}
\usepackage{subcaption}
\usepackage{amssymb}
\usepackage{multirow}
\usepackage{algorithm}
\usepackage{algpseudocode}
\usepackage{url}
\usepackage{bbm}
\usepackage[noabbrev,capitalize]{cleveref}
\newcommand{\RN}[1]{%
  \textup{\uppercase\expandafter{\romannumeral#1}}%
}

\newcommand{\dataset}[1]{\mathbb{X}_{#1}}
\newcommand{\data}[1]{X_{#1}}

\newcommand{\modelname}[1]{M_{#1}}
\newcommand{\modelclf}{M_{\text{CLF}}}

\newcommand{\modelio}[3]{#1\left(#2\right)\left[#3\right]}

\newcommand{\sword}[1]{w_{{#1}}}

\newcommand{\semb}[1]{\vec{e}_{{#1}}}
\newcommand{\interpret}[3]{\phi_{#1}({#2}, {#3})}

\newcommand{\Real}{\mathbb{R}}

\newcommand{\defeq}{\mathrel{\stackrel{\textnormal{\tiny def}}{=}}}

\usepackage[export]{adjustbox}
\usepackage{subcaption}

\usepackage[disable]{todonotes}
\usepackage{array}
\newcolumntype{H}{>{\setbox0=\hbox\bgroup}c<{\egroup}@{}}
\usepackage{calc}
\usepackage{enumitem}

\title{Efficient Shapley Values Estimation by Amortization for Text Classification
}

\author{Chenghao Yang$^{1,4,}\thanks{$^*$Work done during full-time work at AWS AI}$~~, Fan Yin$^{2}$, He He$^{3,4}$, Kai-Wei Chang$^{2,5}$, Xiaofei Ma$^{4}$, Bing Xiang$^{4}$ \\
$^1$ Unversity of Chicago $\qquad$
$^2$ University of California, Los Angeles \\
$^3$ New York University $\qquad$
$^4$ AWS AI Labs $\qquad$ $^5$ Amazon Alexa AI  \\
\texttt{yangalan1996@gmail.com, fanyin20@cs.ucla.edu} \\
\texttt{\{hehea, kaiweic, xiaofeim, bxiang\}@amazon.com}
}

\newcommand{\shortparagraph}[1]{\noindent\textbf{#1}}
\usepackage{array}
    \newcolumntype{P}[1]{>{\centering\arraybackslash}p{#1}}

\begin{document}

\maketitle

\begin{abstract}
Despite the popularity of Shapley Values
in explaining neural text classification models,
computing them is prohibitive for large pretrained models due to a large number of model evaluations.
In practice, Shapley Values are often estimated with a small number of stochastic model evaluations.
However, we show that the estimated Shapley Values are sensitive to random seed choices -- the top-ranked features often have little overlap across different seeds, especially on examples with longer input texts. This can only be mitigated by aggregating thousands of model evaluations, which on the other hand, induces substantial computational overheads.
To mitigate the trade-off between stability and efficiency,
we develop an amortized 
model that directly predicts each input feature's Shapley Value without additional model evaluations.
It is trained on a set of examples whose Shapley Values are estimated from a large number of model evaluations to ensure stability. 
Experimental results on two text classification datasets
demonstrate that  
our amortized model estimates 
Shapley Values accurately
with up to 60 times speedup compared to traditional methods. Furthermore, the estimated values are stable as the inference  is deterministic. We release our code at \url{https://github.com/yangalan123/Amortized-Interpretability}.
\end{abstract}

\section{Introduction}
Many powerful natural language processing (NLP) models used in commercial systems only allow users to access model outputs. When these systems are applied in high-stakes domains, such as healthcare, finance, and law, it is essential to interpret how these models come to their decisions. 
To this end, post-hoc black-box explanation methods have been proposed to identify the input features that are most critical to model predictions~\citep{ribeiro2016lime, lundberg2017unified}. A famous class of post-hoc black-box local explanation methods takes advantage of the Shapley Values~\citep{shapley:book1952} to identify important input features, such as Shapley Value Sampling (SVS)~\citep{strumbelj2010efficient} and KernelSHAP (KS)~\citep{lundberg2017unified}.
These methods typically start by sampling  permutations of the input features (``\emph{perturbation samples}'') and aggregating 
model output changes 
over the perturbation samples. Then, they assign an \emph{explanation score} for each input feature to indicate its contribution to the prediction.
\begin{figure}[t]
\small
\centering
\begin{minipage}{\columnwidth}
\begin{tabular}{P{0.8cm}|c}
\toprule
\textbf{Seed=1} & \adjincludegraphics[width=0.78\columnwidth,valign=c,trim={0 {.38\height} 0 {.05\height}},clip]{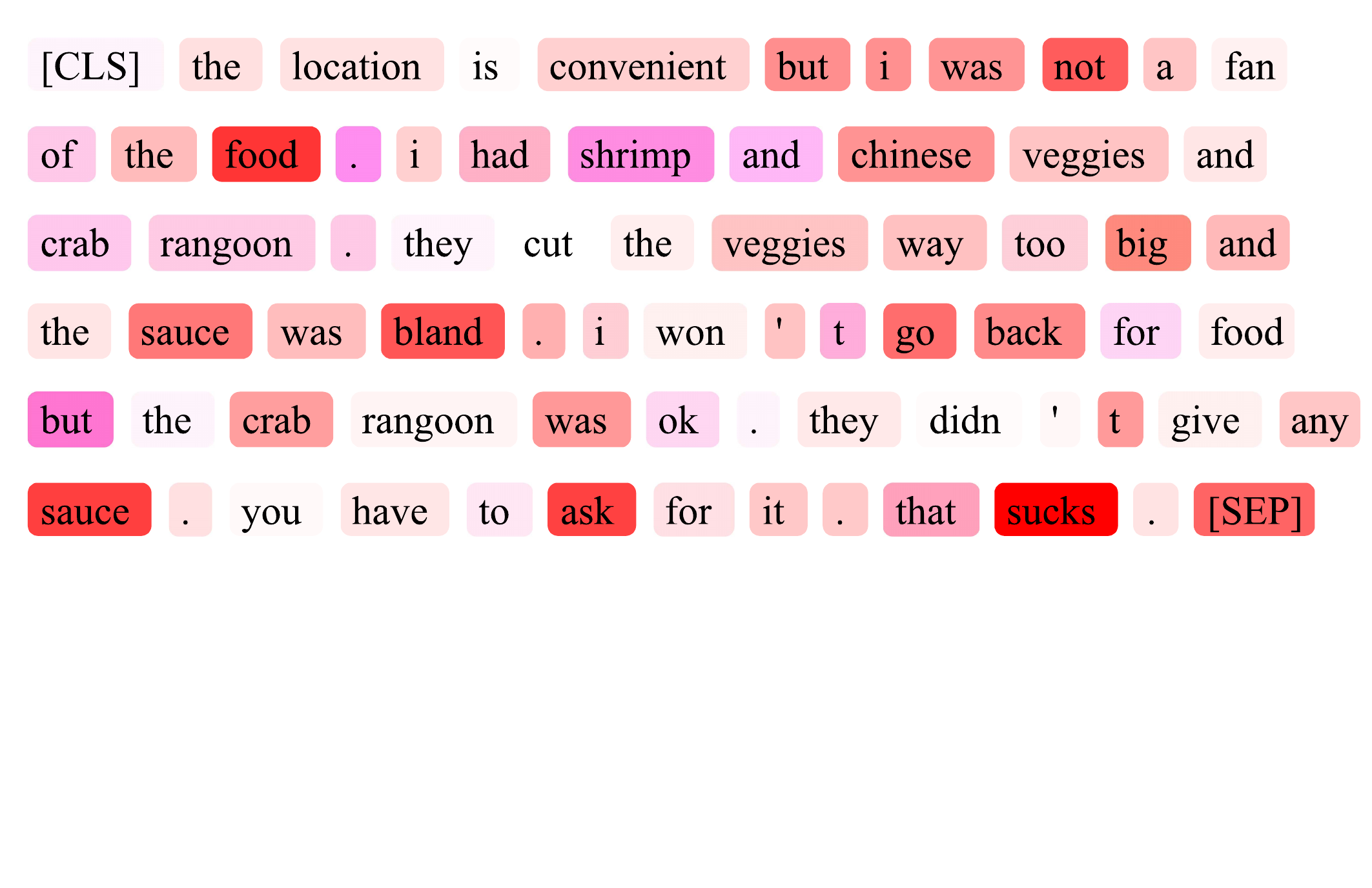} \\
\midrule
\textbf{Seed=2} & \adjincludegraphics[width=0.78\columnwidth,valign=c,trim={0 {.38\height} 0 {.05\height}},clip]{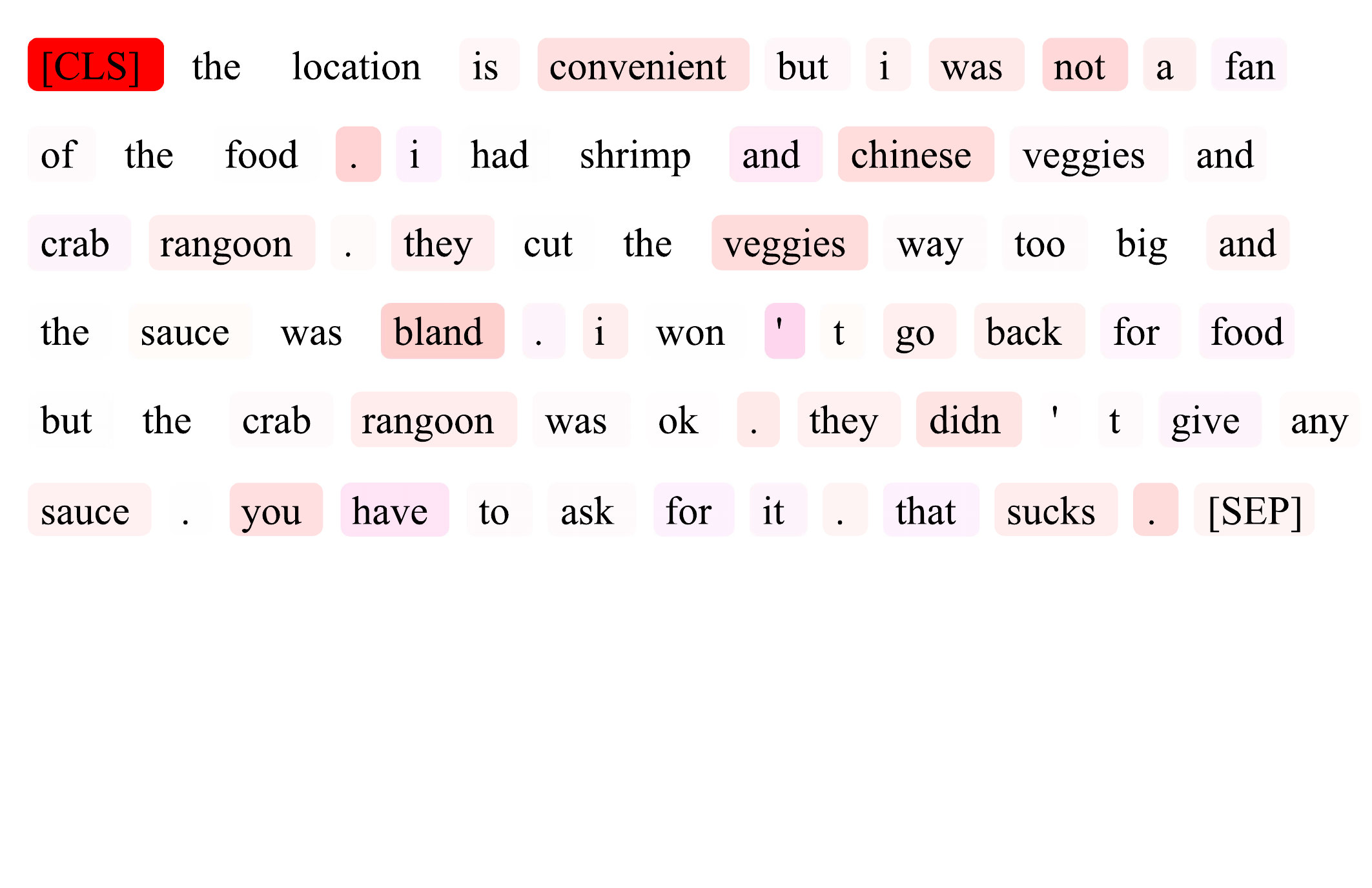} \\
\bottomrule
\end{tabular}
\end{minipage}
\caption{Heatmaps of explanation scores  of an example from Yelp-Polarity based on two runs of KernelSHAP (KS)  using different random seeds. KS is run on a fine-tuned BERT model using $200$ samples per instance (approx. $3.47$s per instance on average using a single A100 GPU, more than $150$ times slower than one forward inference of the BERT model). The darker each token is, the higher its explanation score. Clearly, interpretation results are significantly different when using different seeds.
}
\label{fig: intro_example}
\end{figure}

\begin{figure*}[tbp!]
\centering
\includegraphics[trim={0cm 2cm 0cm 0cm},clip, height=6.21cm]{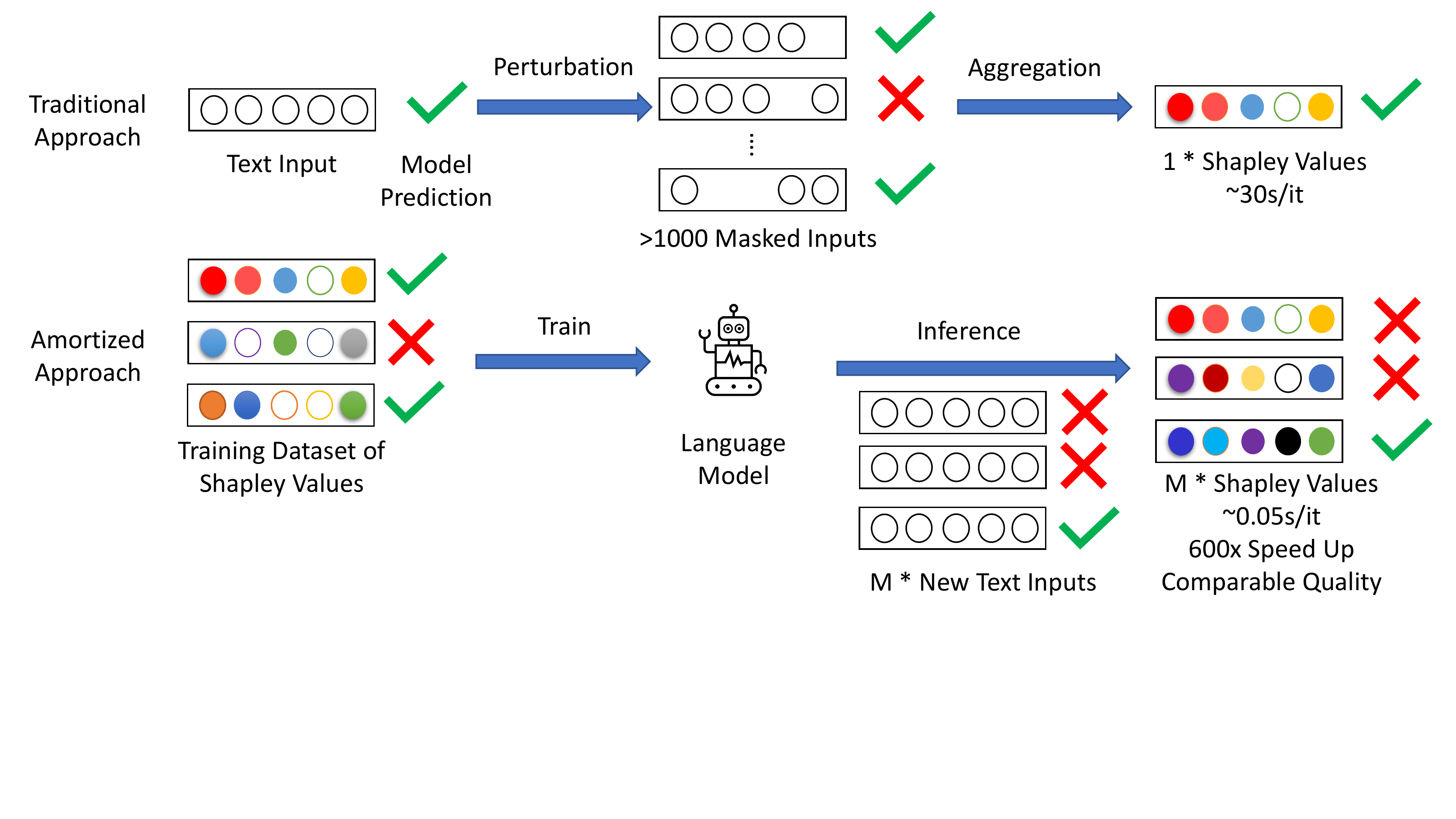}
\vspace{-11mm}
\caption{Illustration of our proposed Amortized Model. Black-outlined circles represent original inputs without Shapley Values, while circles with colored outlines or colored fills denote inputs with Shapley Values.
}
\label{fig: intro_amortized_model}
\end{figure*}

Despite the widespread usage of Shapley Values methods, we observe that when they are applied to text data, the estimated explanation score for each token varies significantly with the random seeds used for sampling.
\cref{fig: intro_example} shows an example of interpreting a BERT-based sentiment classifier~\citep{devlin2019bert} on Yelp-Polarity dataset, a restaurant 
review dataset~\citep{zhangCharacterlevelConvolutionalNetworks2015} by KS. The set of tokens with high explanation scores varies significantly when using different random seeds.
They become stable only when the number of perturbation samples increases to more than 2,000.
As KS requires model prediction for each perturbation sample, the inference cost can be substantial.
For example, it takes about 183 seconds to interpret each instance in Yelp-Polarity 
using the KS Captum implementation~\citep{kokhlikyan2020captum} on an A100 GPU.
In addition, this issue becomes more severe when the input text gets longer, as more perturbation samples are needed for reliable estimation of Shapley Values. This sensitivity to the sampling process leads to an unreliable interpretation of the model predictions and hinders developers from understanding model behavior. 

 To achieve a better trade-off between efficiency and stability, we propose a simple yet effective amortization method to estimate the explanation scores. Motivated by the observation that different instances might share a similar set of important words (e.g., in sentiment classification, emotional words are strong label indicators~\citep{taboada2011lexicon}), an amortized model can leverage similar interpretation patterns across instances when predicting the explanation scores.
Specifically, we amortize the cost of computing explanation scores by precomputing them on a set of training examples and train an amortized model to predict the explanation scores given the input.
At inference time, our amortized model 
directly outputs explanation scores for new instances. Although we need to collect a training set for every model we wish to interpret, our experiments show that with as few as 5000 training instances, the amortized model achieves high estimation accuracy. We show our proposed amortized model in \cref{fig: intro_amortized_model}.%

The experimental results demonstrate the efficiency and effectiveness of our approach. First, our model reduces the computation time from about 3.47s per instance to less than 50ms,\footnote{On Yelp-Polarity dataset and using A100 GPU, we compare with typical KS running with 200 samples.} which is 60 times faster than the baseline methods. 
Second, our model is robust to randomness in training (e.g., random initialization, random seeds used for generating reference explanation scores in the training dataset), and produces stable estimations over different random seeds. Third, we show that the amortized model can be used along with SVS to perform \emph{local adaption}, i.e., adapting to specific instances at inference time, thus further improving performance if more computation is available (\ref{exp: local_adaption}). Finally, we evaluate our model from the functionality perspective~\citep{doshi2017towards, ye2022can} 
by examining the quality 
of the explanation in downstream tasks. We perform case studies on feature selection and domain calibration using the estimated explanation scores, and show that our method outperforms 
the computationally expensive KS method.

\section{Related Works}

\shortparagraph{Post-Hoc Local Explanation Methods}
Post-hoc local
explanations are proposed to understand the prediction process of neural models~\citep{simonyan2013deep, ribeiro2016lime, lundberg2017unified, shrikumar2017learning}. They work by assigning an explanation score to each feature (e.g., a token) in an instance (``local'') to indicate its contribution to the model prediction. In this paper, we focus on studying 
KernelSHAP (KS)~\citep{lundberg2017unified}, an \textit{additive feature attribution method} that estimates the Shapley Value~\citep{shapley:book1952} for each feature. 

There are other interpretability methods in NLP. 
For example, gradient-based methods~\citep{simonyan2013deep, li2016visualizing}, which use the gradient w.r.t. each input dimension as a measure for its saliency. Reference-based methods~\citep{shrikumar2017learning, sundararajan2017axiomatic} consider the model output difference between the original input and reference input (e.g., zero embedding vectors). 

\shortparagraph{Shapley Values Estimation}  Shapley Values are concepts from game theory to attribute total contribution to individual features. However, in practice estimating Shapley values requires prohibitively high cost for computation, especially when explaining the prediction on long documents in NLP. KS works as an efficient way to approximate Shapley Values. Previous work on estimating Shapley Values mainly focuses on accelerating the sampling process~\citep{jethani2021fastshap, covert2021improving,parvez-chang-2021-evaluating,mitchell2022sampling} or removing redundant features~\citep{aas2021explaining, covert2021explaining}. In this work, we propose a new method to combat this challenge by training an amortized model.

\shortparagraph{Robustness of Local Explanation Methods} Despite being widely adopted, there has been a long discussion on the actual quality of explanation methods. Recently, people have found that explanation methods can assign substantially different attributions to similar inputs~\citep{alvarez2018robustness, ghorbani2019interpretation, kindermans2019reliability, yeh2019fidelity, slack2021reliable, yin2022sensitivity}, i.e., they are not robust enough, which adds to the concerns about how faithful these explanations are~\citep{doshi2017towards, Adebayo2018SanityCF, jacovi2020towards}. In addition to previous work focusing on robustness against input perturbations, we demonstrate that even just changing the random seeds can cause the estimated Shapley Values to be weakly-correlated with each other, unless a large number of perturbation samples are used (which incurs high computational cost).

\shortparagraph{Amortized Explanation Methods} Our method is similar to recent works on amortized explanation models including  CXPlain~\citep{schwab2019cxplain} and FastSHAP~\citep{jethani2021fastshap}), where they also aim to improve the computational efficiency of explanation methods. 
The key differences  are: 
1) We do not make causal assumptions between input features and model outputs; and 2) we focus on text domains, where each feature is a discrete token (typical optimization methods for continuous variables do not directly apply). 

\section{Background}
\label{sec:bg}
In this section, we  briefly review the basics of Shapley Values, focusing on its application to the text classification task.

\shortparagraph{Local explanation of black-box text classification models.}
In text classification tasks, inputs are usually sequences of discrete tokens 
$\data{}=[\sword{1}, \sword{2}, \dots, \sword{L}]$. 
Here 
$L$ 
is the length of $\data{}$ and may vary across examples;
$\sword{j}$ 
is the
$j$-th token of $\data{}$.
The classification model  $\modelclf$ takes the input 
$\data{}$ 
and predict the label as 
$\hat{y} = \argmax_{y \in \mathcal{Y}} \modelio{\modelclf}{\data{}}{y}$.
Local explanation methods treat each data instance independently
and compute an explanation score $\interpret{}{j}{y}$,
representing the contribution of $\sword{j}$ to the label $y$.
Usually, we care about the explanation scores when $y=\hat{y}$. 

\shortparagraph{Shapley Values (SV)}
are concepts from game theory originally developed to assign credits in cooperative games~\citep{shapley:book1952, strumbelj2010efficient, lundberg2017unified, covert2021explaining}.
Let $s \in \left\{0, 1\right\}^L$ be a masking of the input and define $\data{s} \defeq \left\{\sword{i} \right\}_{i:s_i=1}$ as the \textit{perturbed input} 
that consists of unmasked tokens $x_i$ (where the corresponding mask $s_i$ has a value of 1). In this paper, we follow the common practice~\citep{ye2021connecting, ye2022can, yin2022sensitivity} to replace masked tokens with \texttt{[PAD]} in the input before sending it to the classifier. 
Let $|s|$ represent the number of non-zero terms in $s$. Shapley Values $\interpret{\text{SV}}{i}{y}$  ~\citep{shapley:book1952} are computed by:
\begin{align}
\small
    &\interpret{\text{SV}}{i}{y} = \frac{1}{L} \sum_{s: s_i \neq 1} {L-1 \choose |s|}^{-1} \nonumber \\
    &\left( \modelio{\modelclf}{\data{s } \cup \left\{\sword{i}\right\} }{y} - \modelio{\modelclf}{\data{s}}{y} \right).
\end{align}
Intuitively, 
$\interpret{\text{SV}}{i}{y}$ computes the marginal contributions of each token 
to the model prediction.

Computing SV is known to be NP-hard~\citep{deng1994complexity}.
In practice, we estimate Shapley Values approximately for efficiency. 
Shapley Values Sampling (SVS)~\citep{castro2009polynomial, strumbelj2010efficient} is a widely-used Monte-Carlo estimator of SV: 

{\small
\begin{align}
\small
        &\interpret{\text{SVS}}{i}{y} = \frac{1}{m} \sum_{\substack{\sigma_j \in \Pi(L)\\ 1\leq j\leq m}}\sum_{i \in \sigma_j} 
        \nonumber \\ 
        &\left[ \modelio{\modelclf}{\data{\mathbb{S}\left([\sigma_j]_{i-1} \cup \left\{i\right\}\right)}}{y}  \right. 
        \left. -\modelio{\modelclf}{\data{\mathbb{S}\left([\sigma_j]_{i-1}\right)}}{y} \right].  \label{eq: mc_shapley} 
\end{align}
}%
Here $\sigma_j \in \Pi(L)$ is the sampled \textbf{ordering} and 
 $[\sigma_j]$ is the non-ordered \textbf{set} of indices for $\sigma_j$. 
$[\sigma_j]_{i-1}$ represents the \textbf{set}  of indices ranked lower than $i$ in $\sigma_j$ . $\mathbb{S}([\sigma_j])$ maps the indices \text{set} $[\sigma_j]$
to a mask $s \in \{0, 1\}^L$ such that $s_i=\mathbf{1}[i \in [\sigma_j]]$. $m$ is the number of \emph{perturbation samples} used for computing SVS. 

\shortparagraph{KernelSHAP}
Although SVS has successfully reduced the exponential time complexity to polynomial, it still requires sampling permutations and needs to do sequential updates 
following sampled orderings and computing the explanation scores,
which is an apparent efficiency bottleneck. 
\citet{lundberg2017unified} introduce a more efficient estimator, KernelSHAP (KS), which allows better parallelism 
and computing explanation scores for all tokens at once using linear regression. That is achieved by showing that computing SV is equivalent to solving the following optimization problem:
\begin{align}
     &\interpret{\text{KS}}{\cdot}{y} \approx \argmin_{\interpret{}{\cdot}{y}} \frac{1}{m}  \nonumber \\
     &\sum_{\substack{s(k) \sim p(s)\\1\leq k\leq m}}[\modelio{\modelclf}{\data{s(k)}}{y}  - \vec{s}(k)^T\interpret{}{\cdot}{y}]^2 \label{eq: kernelshap-approx}, \\
     &\textrm{s.t.} \quad  \mathbf{1}^T\interpret{}{\cdot}{y} = \modelio{\modelclf}{\data{}}{y} - \modelio{\modelclf}{\varnothing}{y}, \nonumber
     \label{eq: sufficiency constraints}
\end{align}
where $\vec{s}(k)$ is the one-hot vector
corresponding to the mask\footnote{Note, $s(k)$ is the $k$-th \textbf{mask sample} while $s_i \in \{0, 1\}$ is the $i$-th dimension of the \textbf{mask sample} $s$.} $s(k)$
sampled from the Shapley Kernel  $p(s) = \frac{L-1}{{L \choose |s|}|s|(L-|s|)}$.
$m$ is again the number of perturbation samples. 
We will use ``SVS-$m$'' and ``KS-$m$'' in the rest of the paper to indicate the sample size for SVS and KS. 
In practice, the specific perturbation samples depend on the random seed of the sampler, and we will show that the explanation scores are highly sensitive to the random seed under a small sample size. 

Note that the larger the number of perturbation samples, the more model evaluations are required for a single instance, which can be computationally expensive for large Transformer models. Therefore, the main performance bottleneck is the number of model evaluations. 

\section{Stability of Local Explanation}
\label{sec: stability}
One of the most common applications of SV is feature selection, which selects the most important features by following the order of the explanation scores. People commonly use KS with an affordable number of perturbation samples in practice (the typical numbers of perturbation samples used 
in the literature
are around $25$, $200$, $2000$). 
 However, as we see in \cref{fig: intro_example}, the ranking of the scores can be quite sensitive to random seeds when using stochastic estimation of SV. 
In this section, we investigate this stability issue. We demonstrate stochastic approximation of SV is unstable 
in text 
classification tasks under common settings, especially with long texts. In particular, when ranking input tokens based on explanation scores, Spearman's correlation between rankings across different runs is low. 

\shortparagraph{Measuring ranking stability.} 
Given explanation scores produced by different random seeds using an SV estimator, we want to measure the difference between these scores. Specifically, we are interested in the difference in the rankings of the scores as this is what we use for feature selection. To measure the ranking stability of multiple runs using different random seeds,
we compute Spearman's correlation between any two of them and use the average Spearman's correlation as the measure of the ranking stability. 
In addition, we follow~\citet{ghorbani2019interpretation} to report Top-K 
intersections
between two rankings, since in many applications only the top features are of explanatory interest. We measure the size of the intersection 
of Top-K features from two different runs.
\begin{table*}[!ht]
\small
\centering
\begin{tabular}{lccccr}
\hline
Setting & Spearman & Top-5 Inter. & Top-10 Inter. & MSE & Running Time \\ \hline
SVS-25 & $0.84 (\pm 0.00)$ & $3.41 (\pm 0.00)$ & $7.02 (\pm 0.00)$ & $0.01 (\pm 0.00)$  & $183.72$s/it \\
KS-25 & $0.04 (\pm 0.00)$ & $0.43 (\pm 0.01)$ & $1.45 (\pm 0.01)$ & $0.00 (\pm 0.00)$ & $1.92$s/it \\
KS-200 & $0.16 (\pm 0.00)$ & $1.09 (\pm 0.01)$ & $2.47 (\pm 0.00)$ & $0.82 (\pm 0.29)$ & $3.47$s/it \\
KS-2000 & $0.37 (\pm 0.00)$ & $2.45 (\pm 0.01)$ & $4.38 (\pm 0.05)$ & $0.03 (\pm 0.00)$ & $33.40$s/it \\
KS-8000 & $0.63 (\pm 0.00)$ & $3.73 (\pm 0.02)$ & $6.93 (\pm 0.01)$ & $0.01 (\pm 0.00)$ & $123.29$s/it \\ 
\hline
\end{tabular}%
\caption{Ranking stability experiments on the Yelp-Polarity dataset. Each local explanation setting is evaluated across 5  runs with different random seeds. ``Top-K Inter.'' denotes top-K intersection. 
All values in this table are absolute values. Here we can see a clear trade-off between stability and computation cost. 
}
\label{tab:ranking_instability_yelp}
\end{table*}
\begin{table*}[]
\small
\centering
\begin{tabular}{lccccr}
\hline
Setting & Spearman & Top-5 Inter. & Top-10 Inter. & MSE & Running Time \\ \hline
SVS-25 & $0.75 (\pm 0.00)$ & $3.54 (\pm 0.02)$ & $7.46 (\pm 0.02)$ & $0.02 (\pm 0.00)$ & $128.07$s/it \\
KS-25 & $0.06 (\pm 0.00)$ & $0.97 (\pm 0.01)$ & $3.41 (\pm 0.03)$ & $0.01 (\pm 0.00)$ & $0.33$s/it \\
KS-200 & $0.24 (\pm 0.00)$ & $1.79 (\pm 0.01)$ & $4.37 (\pm 0.03)$ & $0.07 (\pm 0.00)$ & $2.04$s/it \\
KS-2000 & $0.52 (\pm 0.00)$ & $3.19 (\pm 0.00)$ & $6.09 (\pm 0.00)$ & $0.03 (\pm 0.00)$ & $20.39$s/it \\
KS-8000 & $0.76 (\pm 0.00)$ & $4.08 (\pm 0.02)$ & $7.74 (\pm 0.02)$ & $0.01 (\pm 0.00)$ & $89.48$s/it \\ \hline
\end{tabular}%
\caption{Ranking stability experiments on the MNLI dataset. 
}
\label{tab:ranking_instability_mnli}
\end{table*}

\shortparagraph{Setup.} We conduct our experiments on the validation set of the Yelp-Polarity dataset~\citep{zhangCharacterlevelConvolutionalNetworks2015} and MNLI dataset~\citep{williams2018broad}. Yelp-Polarity is a binary sentiment classification task and MNLI is a three-way textual entailment classification task. 
We conduct experiments on $500$ random samples with balanced labels (we refer to these datasets as ``Stability Evaluation Sets'' subsequently). Results are averaged over $5$ different random seeds.\footnote{We take more than 2,000 hours on a single A100 GPU for all experiments in this section.} We use the publicly available fine-tuned BERT-base-uncased checkpoints\footnote{Yelp-Polarity: \url{https://huggingface.co/textattack/bert-base-uncased-yelp-polarity}\\MNLI: \url{https://huggingface.co/textattack/bert-base-uncased-MNLI}}~\citep{morris2020textattack} as the target models to interpret and use the implementation of Captum~\citep{kokhlikyan2020captum} to compute the explanation scores for both KS and SVS. 
For each explanation method, we test with the recommended numbers of perturbation samples\footnote{For SVS, the recommended number of perturbation samples is $25$ in Captum. For KS, to our best knowledge, the typical numbers of perturbation samples used in previous works are $25, 200, 2000$. We also include KS-$8000$ to see how stable KS can be given much longer running time.}
 used to compute the explanation scores for every instance. For Top-K intersections, we report results with  $K=5$ and $K=10$.

\shortparagraph{Trade-off between stability and computation cost.} The ranking stability results are listed in \cref{tab:ranking_instability_yelp} 
and \cref{tab:ranking_instability_mnli} 
for Yelp-Polarity and MNLI
datasets.
We observe that using 25 to 200 perturbation samples, the stability of the explanation scores is low (Spearman's correlation is only 0.16).
Sampling more perturbed inputs 
makes the scores more stable. However, the computational cost explodes at the same time, going from one second to two minutes per instance. 
To reduce the sensitivity to an acceptable level (i.e., making the Spearman's correlation between two different runs above $0.40$, which indicates moderate  correlation~\cite{akoglu2018user}), we usually need thousands of model evaluations and spend roughly $33.40$ seconds per instance. 

\shortparagraph{Low MSE does not imply stability.} Mean Squared Error (MSE) is commonly used to evaluate the distance between two lists of 
explanation scores. 
In \cref{tab:ranking_instability_yelp}, we observe that MSE only weakly correlates with ranking stability (e.g., For Yelp-Polarity, $R=-0.41$ and $p<0.05$, so the correlation is not significant). Even when the difference of MSE for different settings is as low as 0.01, the correlation between rankings produced by explanations can still be low. Therefore, from users' perspectives, low MSEs do not mean the explanations are reliable as they can suggest distinct rankings. 

\begin{figure*}[tbp!]
\centering
\begin{subfigure}[b]{0.45\textwidth}
\centering
\includegraphics[ height=4cm]{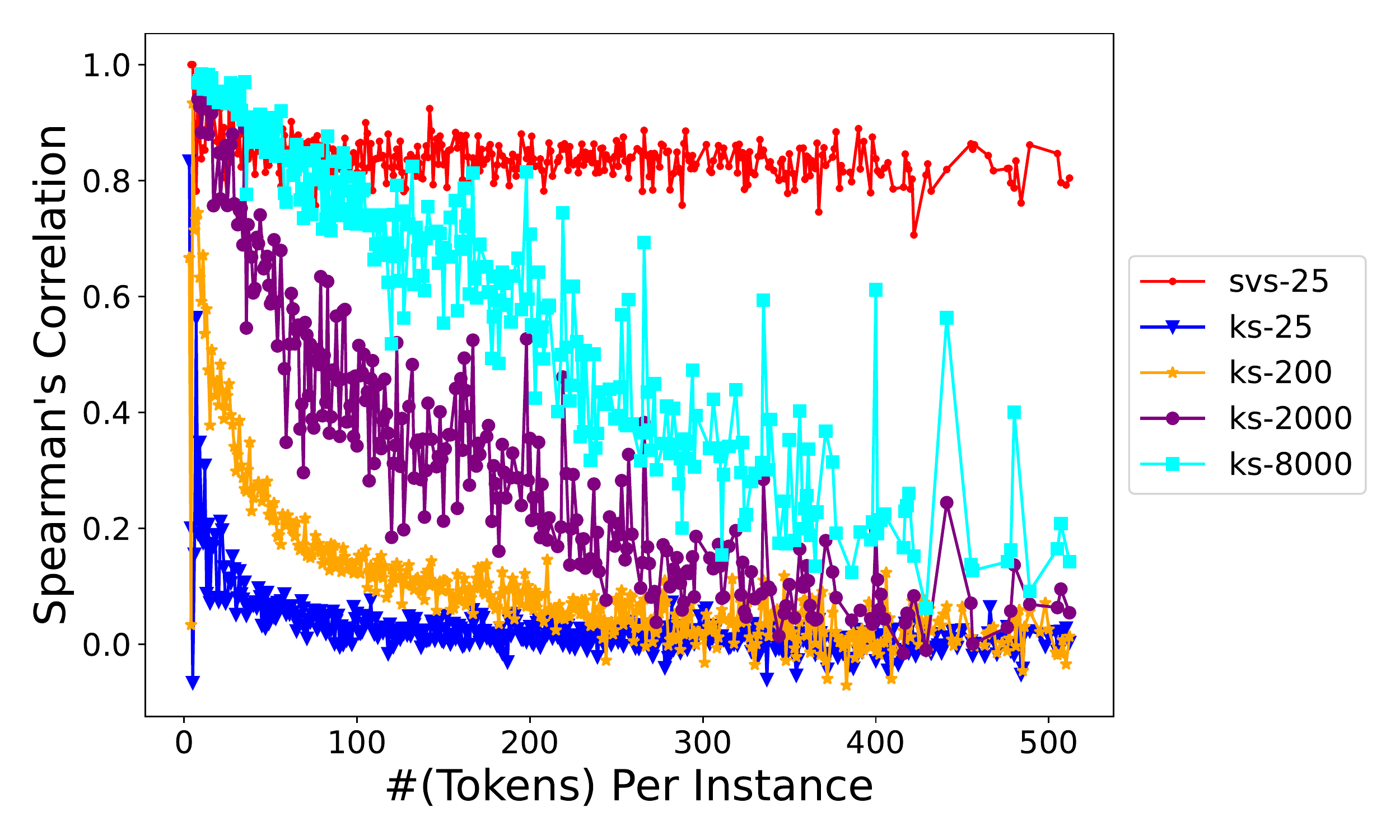}
\vspace{-3mm}
\caption{Yelp-Polarity
}
\label{fig: internal_correlation_with_length_yelp}
\end{subfigure}
\begin{subfigure}[b]{0.45\textwidth}
\small
\centering
\includegraphics[ height=4cm]{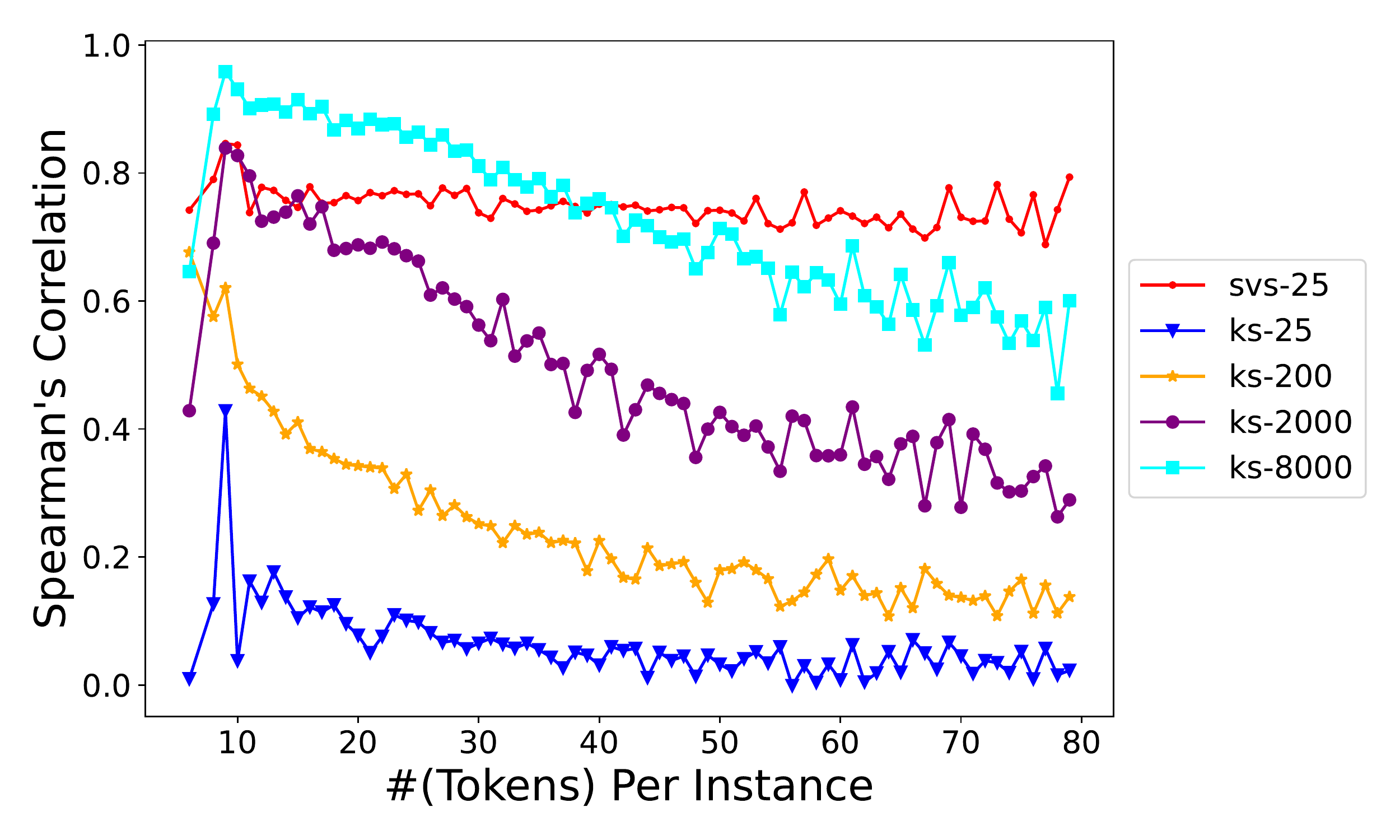}
\vspace{-3mm}
\caption{MNLI
}
\label{fig: internal_correlation_with_length_mnli}
\end{subfigure}
\caption{Ranking stability over different input lengths on Yelp-Polarity and MNLI datasets. We observe that longer input suffers more from instability. 
}
\label{fig: internal_correlation_with_length}
\end{figure*}

\shortparagraph{Longer input suffers more from instability.} We also plot the Spearman's correlation decomposed at different input lengths in \cref{fig: internal_correlation_with_length}.
Here, we observe a clear trend that the ranking stability degrades significantly even at an input length of 20 tokens. The general trend is that the longer the input length is, the worse the ranking stability. The same trend holds across datasets. As many NLP tasks involve sentences longer than 20 tokens (e.g., SST-2~\cite{socher-etal-2013-recursive}, MNLI~\cite{williams2018broad}), obtaining stable explanations to analyze NLP models can be quite challenging. 

\shortparagraph{Discussion: why Shapley Values estimation is unstable in text domain?}
One of the most prominent characteristics of the text domain is that individual tokens/n-grams can have a large impact on the label. Thus they need to be all included in the perturbation samples for an accurate estimate.
When the input length grows, the number of n-grams will grow fast. As shown in \cref{sec:bg}, the probability of certain n-grams getting sampled is drastically reduced as each n-gram will be sampled with equivalent probability. Therefore, the observed model output will have a large variance as certain n-grams may not get sampled. A concurrent work \cite{kwon2022weightedshap} 
presented a related theoretical analysis on why the uniform sampling setting in SV computation can lead to suboptimal attribution. 

\section{Amortized Inference for Shapley Values
}
\label{sec: method}
Motivated by the above observation, 
we propose to train an amortized model to predict the explanation scores given an input \emph{without any model evaluation on perturbation samples}. The inference cost is thus amortized by training on a set of pre-computed reliable explanation scores. 

We build an amortized explanation model for text classification in two stages. In the first stage, we construct a training set for the amortized model. 
We compute reliable explanation scores as the reference scores for training using the existing SV estimator. As shown in \cref{sec: stability}, SVS-25 is the most stable SV estimator and we use it to obtain reference scores. 
In the second stage, we train a BERT-based amortized model that takes the text as input and outputs the explanation scores using MSE loss. 

Specifically, given input tokens $\data{}$, we  use a pretrained language model $\modelname{\mathrm{LM}}$
to encode words into $d$-dim embeddings $\vec{e} =\modelname{\mathrm{LM}}(\data{})= [\semb{1}, \dots, \semb{L(\data{})}] \in \Real^{L(\data{}) \times d}$. Then, we use a linear layer to transform each $\semb{i}$ to the predicted explanation score $\interpret{AM}{i}{\hat{y}_i} = W\semb{i} + b$. 
To train the model, we use MSE loss to fit $\interpret{AM}{i}{\hat{y}}$ to the pre-computed reference scores $\interpret{}{i}{\hat{y}}$ 
over the training set $\dataset{\text{Train}}$. This is an amortized model in the sense that there are no individual sampling and model queries for each test example $\data{}$ as in SVS and KS. When a new sample comes in, the amortized model makes a single inference on the input tokens to predict their explanation scores.  

\begin{algorithm}
\small
\caption{Local Adaption}\label{alg:meta}
\begin{algorithmic}
\Require $m$: the desired number of local adaption perturbation samples, $\modelname{\mathrm{AM}}$: the trained amortized explanation model, $\data{}$: the target data instance that has length $L$, $\hat{y}$: the predicted label, $\modelname{\mathrm{CLF}}$: the target model
\State $\phi \gets \modelname{\mathrm{AM}}(\data{})$
\For{ $j=1$ to $m$}
    \State sample ordering $\sigma$ from permutation $\Pi(L)$
    \State $\phi \gets \phi + \sum_{i}\left[ \modelio{\modelclf}{\data{\mathbb{S}\left([\sigma]_{i-1} \cup \left\{i\right\}\right)}}{\hat{y}} \right.$ \\
    $\left.\quad\quad\qquad\qquad\qquad - \modelio{\modelclf}{\data{\mathbb{S}\left([\sigma]_{i-1}\right)}}{\hat{y}} \right]$%
\EndFor
\State $\phi \gets \frac{\phi}{m}$
\end{algorithmic}
\end{algorithm}

\subsection{Better Fit via Local Adaption}
\label{method: local_adaption}
By amortization, our model can learn to capture the shared feature attribution patterns across data to achieve a good efficiency-stability trade-off. 
We further show that the explanations generated by our amortized model can be used to initialize the explanation scores of SVS. This way, the evaluation of SVS can be significantly sped up compared with using random initialization. On the other hand, applying SVS upon amortized method improves the latter's performance as some important tokens  might not be captured by the amortized method but can be identified by SVS through additional sampling (e.g., low-frequency tokens). 
The detailed algorithm is shown in \cref{alg:meta}. Note that here we can recover the original SVS computation~\citep{strumbelj2010efficient} by replacing $\phi \gets \modelname{\mathrm{AM}}(\data{})$ to be $\phi \gets 0$. $\modelname{\mathrm{AM}}$ is the amortized model trained using MSE as explained earlier. 

\section{Experiments}
\label{sec: experiment}
In this section, we present experiments to demonstrate the properties of the proposed approach in terms of accuracy
against reference scores (\ref{sec:expapp}) and sensitivity to training-time randomness (\ref{sec: training_sensitivity}). We also show that we achieve a better fit via a local adaption method that combines our approach with SVS (\ref{exp: local_adaption}).
Then, we evaluate the quality of the explanations generated by our amortized model on two downstream applications (\ref{sec: downstream_app}). 

\shortparagraph{Setup.}
We conduct experiments on the validation set of Yelp-Polarity and MNLI datasets.
To generate reference explanation scores, we leverage the Thermostat~\citep{feldhus2021thermostat} dataset, which contains 9,815 pre-computed explanation scores of SVS-25 on MNLI. We also
compute explanation scores of SVS-25 for 25,000 instances on Yelp-Polarity. 
We use BERT-base-uncased~\citep{devlin2019bert} for $\modelname{\mathrm{LM}}$. 
For dataset preprocessing and other experiment details, we refer readers to \cref{app: training_details}.

To our best knowledge, FastSHAP~\citep{jethani2021fastshap} is the most relevant work to us that also takes an amortization approach to estimate SV on tabular or image data. We adapt it to explain the text classifier and use it as a baseline to compare with our approach. 
We find it non-trivial to adapt FastSHAP to the text domain.  As pre-trained language models occupy a large amount of GPU memory, we can only use a small batch size with limited perturbation samples (i.e., $32$ perturbation samples per instance). 
This is equivalent to approximate KS-32 and the corresponding reference explanation scores computed by FastSHAP are unstable. 
More details can be found in \cref{app: fastshap}.

\begin{table}[!htbp]
\centering
\resizebox{\columnwidth}{!}{  
\begin{tabular}{@{}ccccc@{}}
\toprule
\multirow{2}{*}{Method} & \multicolumn{2}{c}{MNLI} & \multicolumn{2}{c}{Yelp-Polarity} \\
         & Spearman  & MSE  & Spearman  & MSE   \\ \midrule        
SVS-25          & 0.75                 & 1.90e-2  & 0.84                 & 6.64e-3       \\
KS-25   & 0.17                 & 9.95e-2  & 0.12                 & 4.34e-2       \\
KS-200  & 0.35                 & 7.73e-2  & 0.24                 &  5.77e-2              \\
KS-2000 & 0.60                 & 2.54e-2   & 0.51                 & 1.86e-2      \\
KS-8000 & \textbf{0.74}                 & \textbf{1.25e-2}   & \textbf{0.70}                 & \textbf{6.25e-3}     \\
\midrule
FastSHAP & 0.23 & 1.90e-1  & {0.18} & {7.91e-3} \\
Our Amortized Model & \textbf{0.42}                 & \textbf{9.59e-3} & \textbf{0.61}                 & \textbf{4.46e-6}        \\ 

\bottomrule
\end{tabular}
}
\caption{Spearman's correlation and MSE of variants of SV methods against SVS-25, a proxy of exact SV on MNLI and Yelp-Polarity. As we show in \cref{sec: stability}, MSE correlates poorly with ranking stability and Spearman's correlation should be considered as \textbf{the main metric}. We only list MSE for reference. Bold-faced numbers are the best in each column. Results are averaged over 5 runs. Our amortized model achieves better approximation compared to KS-200 and FastSHAP baseline, but not as good as much more time-consuming methods KS-2000/8000. 
SVS-25 
is listed 
as an upper bound. }
\label{tab:approximation}
\end{table}

\subsection{Shapley Values Approximation}
\label{sec:expapp}
To examine how well our model fits the pre-computed SV (SVS-25), we compute both Spearman's correlation and MSE over the test set. As it is intractable to compute exact Shapley Values for ground truth, we use SVS-25 as a proxy. We also include different settings for KS results over the same test set. KS is also an approximation to permutation-based SV computation~\citep{lundberg2017unified}. 
\cref{tab:approximation} shows 
the correlation and MSE of aforementioned methods against SVS-25.

First, we find that despite the simplicity of our amortized model, the proposed amortized models achieve a high correlation with the reference scores ($0.61 > 0.60$) on Yelp-Polarity.
The correlation between outputs from the amortized models and references is moderate ($0.42 > 0.40$) on MNLI when data size is limited. During inference time, our amortized models output explanation scores for each instance within  $50$ milliseconds, which is about $40$-$60$ times faster than KS-200 and $400$-$600$ times faster than KS-2000 on Yelp-Polarity and MNLI. Although the approximation results are not as good as KS-2000/8000 (which requires far more model evaluations), our approach achieves reasonably good results with orders of magnitude less compute. 

We also find that the amortized model achieves the best
MSE score among all approximation methods. Note that the two metrics, Spearman's correlation and MSE, do not convey the same information. MSE measures how well the reference explanation scores are fitted while Spearman's correlation reflects how well the ranking information is learned. We advocate for reporting both metrics. 

\begin{figure}[tbp!]
  \centering
  \begin{subfigure}[b]{0.48\columnwidth}
  \centering
    \includegraphics[width=\textwidth]{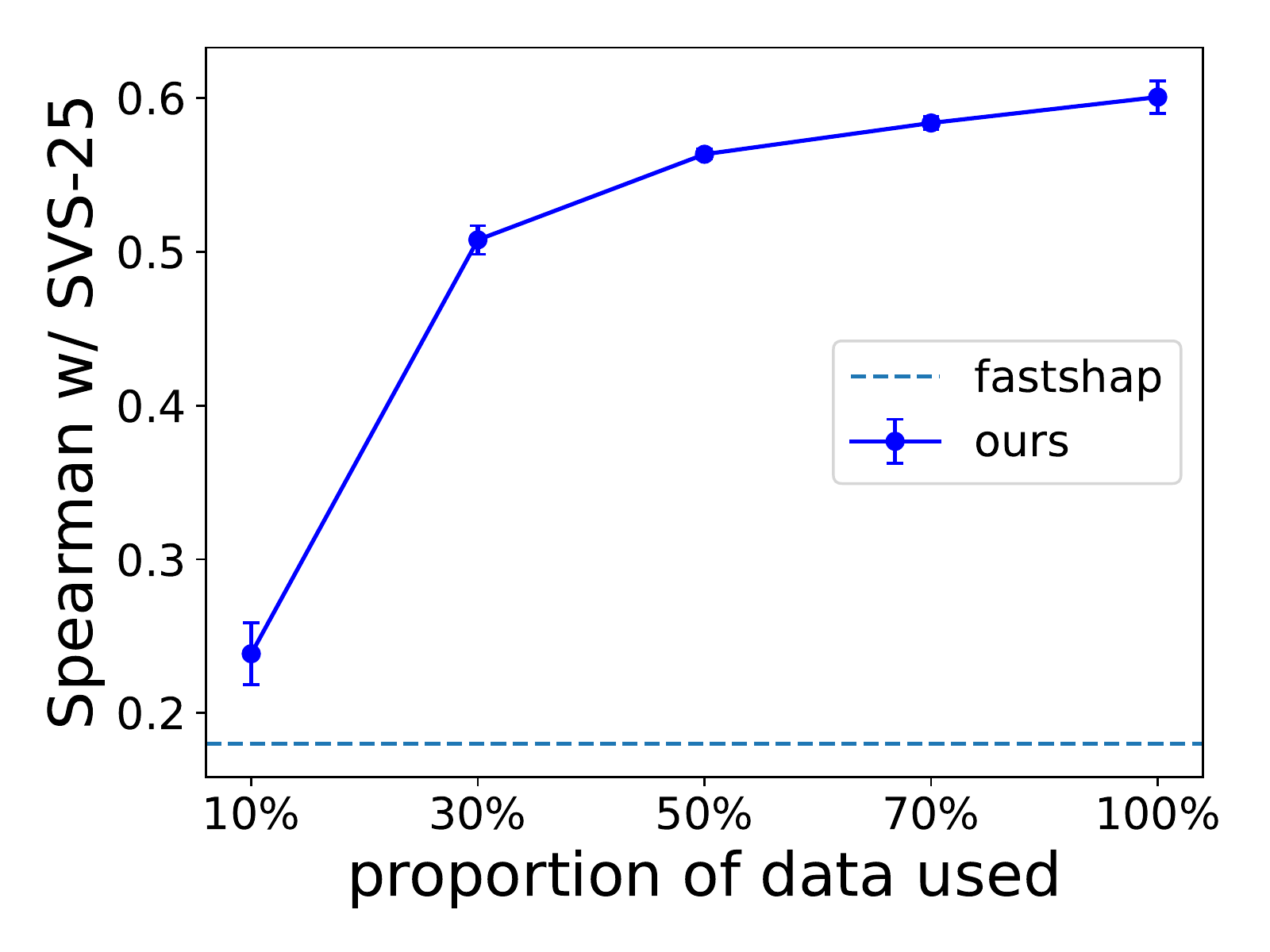}
    \vspace{-7mm}
    \caption{Yelp-Polarity}
    \label{fig:yelp_learning_curve}
  \end{subfigure}
  \begin{subfigure}[b]{0.48\columnwidth}
  \centering
    \includegraphics[width=\textwidth]{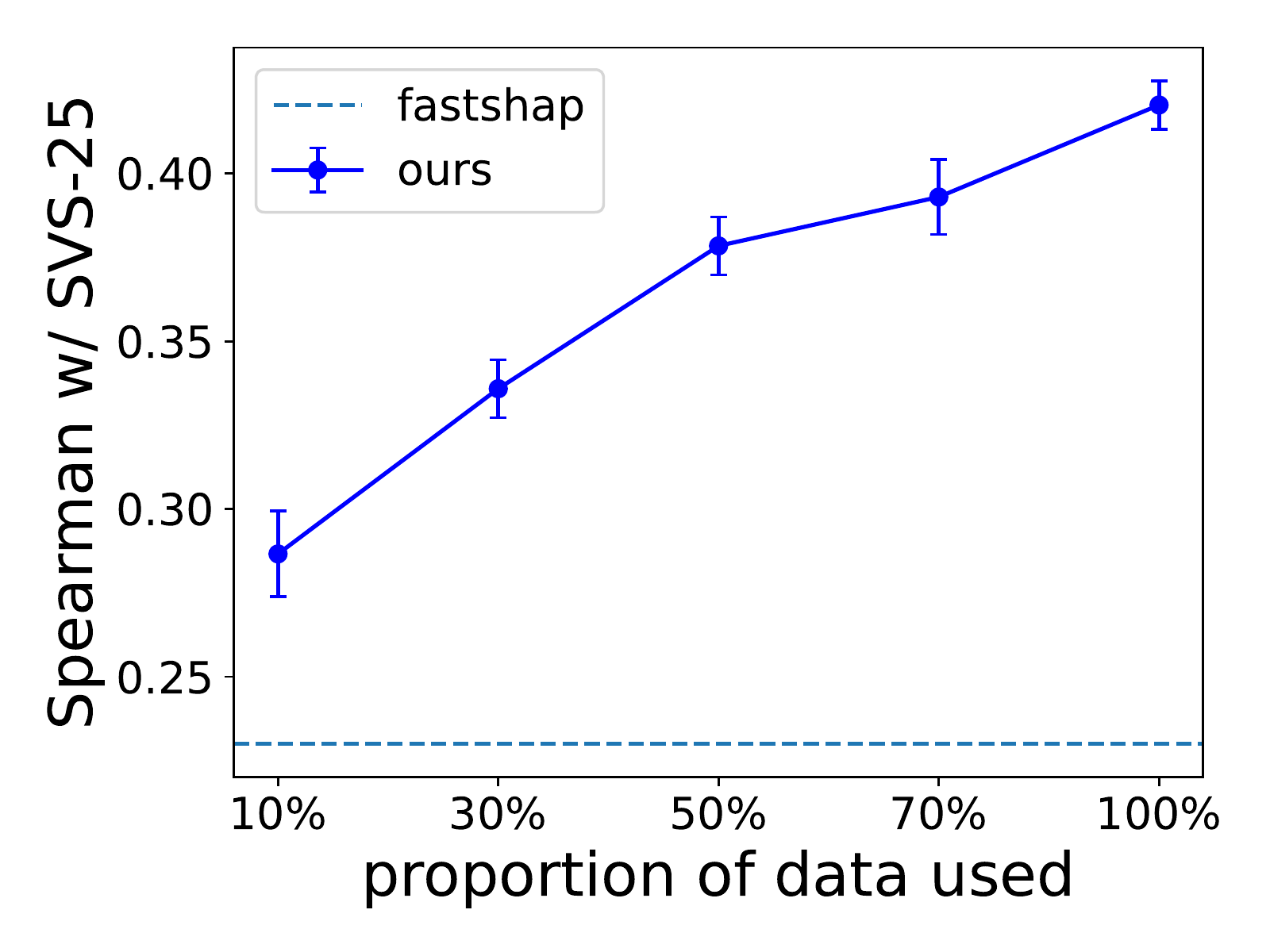}
    \vspace{-7mm}
    \caption{MNLI}
   \label{fig:mnli_learning_curve}
  \end{subfigure}
  \vspace{-3mm}
  \caption{Learning curves for the amortized model over Yelp-Polarity and MNLI datasets. The Spearman's correlations in this figure are computed against SVS-25. We can see our amortized model can learn efficiently even if there is only 10\% data used for training. }
  \label{fig: learning curve}
\end{figure}

\shortparagraph{Cost of training the amortized models}
To produce the training set, we need to pre-compute the explanation scores on a set of data. Although this is a one time cost (for each model), one might wonder how time consuming this step is as we need to run the standard sample-based estimation. As the learning curve shows in \cref{fig: learning curve}, 
we observe that the model achieves good performance with about $25\%$ ($\approx 5,000$ on Yelp-Polarity) instances. 
Additionally, in \cref{sec: domain-transfer}, we  show this one-time training will result in a model transferable to other domains, so we may not need to train a new amortized model for each new domain. 

\begin{table}[!ht]
\centering
\resizebox{\columnwidth}{!}{
\centering
\begin{tabular}{@{}ccc@{}}
\toprule
Training Data Proportion          & Spearman (MNLI) & Spearman (Yelp-Polarity)   \\ \midrule
10\%          & 0.45   & 0.40                     \\
30\%   & 0.57    & 0.65                     \\
50\%  & 0.65  & 0.71                              \\
70\% & 0.65   & 0.72                       \\
100\% & 0.77  & 0.76                        \\ \bottomrule
\end{tabular}
}
\caption{Training time sensitivity study. To evaluate how much the amortized model will be influenced by randomness during training, we sample training data 5 times with different random seeds and then compute the averaged Spearman's correlation among all pairs of runs. 
The standard deviation is less than 1e-2.
Our amortized model is stable against training time randomness with only 10\% of data. 
}
\label{tab: training_time_sensitivity}
\vspace{-3mm}
\end{table}

\subsection{Sensitivity Analysis}
\label{sec: training_sensitivity}
Given a trained amortized model,
there is no randomness when generating explanation scores.
However, there is still some randomness in the training process, including the training data, the random initialization of the output layer and randomness during update such as dropout. 
Therefore, similar to 
\cref{sec: stability}, we study the sensitivity of the amortized model.
 \cref{tab: training_time_sensitivity} 
 shows the results with different training data and random seeds.
We observe that: 1) when using the same data (100\%), random initialization does not affect the outputs of amortized models -- the correlation between different runs is high (i.e., $0.77$ on MNLI and $0.76$ on Yelp-Polarity). 2) With more training samples, the model is more stable. 

\begin{table}[!ht]
\small
\centering
\resizebox{\columnwidth}{!}{
\begin{tabular}{@{}ccc@{}}
\toprule
\multirow{2}{*}{Method} & \multicolumn{1}{c}{MNLI} & \multicolumn{1}{c}{Yelp-Polarity} \\
         & Spearman    & Spearman     \\ \midrule       
SVS-2          & 0.41                   & 0.52                      \\
SVS-3          & 0.47                   & 0.60                      \\
SVS-5          & 0.55                   & 0.69                      \\
SVS-25          & \textbf{0.75}                   & \textbf{0.84}                      \\
\midrule
Our Amortized Model & {0.42}                  & {0.61}                        \\ 
Our Amortized Model (Adapt-2) & {0.47}                  & {0.64}                        \\ 
Our Amortized Model (Adapt-3) & {0.53}                  & {0.69}                        \\ 
Our Amortized Model (Adapt-5) & \textbf{0.57}                  & \textbf{0.71}                        \\ 

\bottomrule
\end{tabular}}
\caption{Approximation results for the Shapley explanation methods on MNLI and Yelp-Polarity datasets. Bold-faced numbers are the best in each column. Results are averaged over 5 runs. Spearman's correlations are computed against SVS-25. Adapt-m means here how many sampled ordering $\sigma$s we used here to do local adaption ($m$ in \cref{alg:meta}). 
}
\label{tab:meta_approximation}

\end{table}

\subsection{Local Adaption}
\label{exp: local_adaption}

The experiment results for Local Adaption (\cref{method: local_adaption}) are shown in \cref{tab:meta_approximation}. Here we can see that: 1) by doing local adaption, we can further improve the approximation results using our amortized model, 2) by using our amortized model as initialization, we can improve the sample efficiency of SVS significantly (by comparing the performance of SVS-X and Adapt-X). These findings hold across datasets.

\subsection{Domain Transferability}
\label{sec: domain-transfer}
To see how well our model performs on out-of-domain data, we train a classification model and its amortized explanation model on Yelp-Polarity and then explain its performance on SST-2~\cite{socher-etal-2013-recursive} validation set. Both tasks are two-way sentiment classification and have significant domain differences. 

Our amortized model achieves a Spearman's correlation of approximately 0.50 with ground truth SV (SVS-25) while only requiring 0.017s per instance.  In comparison, KS-100 achieves a lower Spearman's correlation of 0.46 with the ground truth  and takes 1.6s per instance; KS-200 performs slightly better in Spearman's correlation but requires significantly more time. Thus, our amortized model is more than 90 times faster and more correlated with ground truth Shapley Values. This shows that, once trained, our amortized model can provide efficient and stable estimations of SV even for out-of-domain data. 

In practice, we do not recommend directly explaining model predictions on out-of-domain data without verification, because it may be misaligned with user expectations for explanations, and the out-of-domain explanations may not be reliable~\cite{hase2021out,denain2022auditing}. More exploration on this direction is required but is orthogonal to this work.

\begin{figure*}[tbp!]
  \centering
  \begin{subfigure}[b]{0.48\textwidth}
    \centering
    \includegraphics[width=0.9\textwidth]{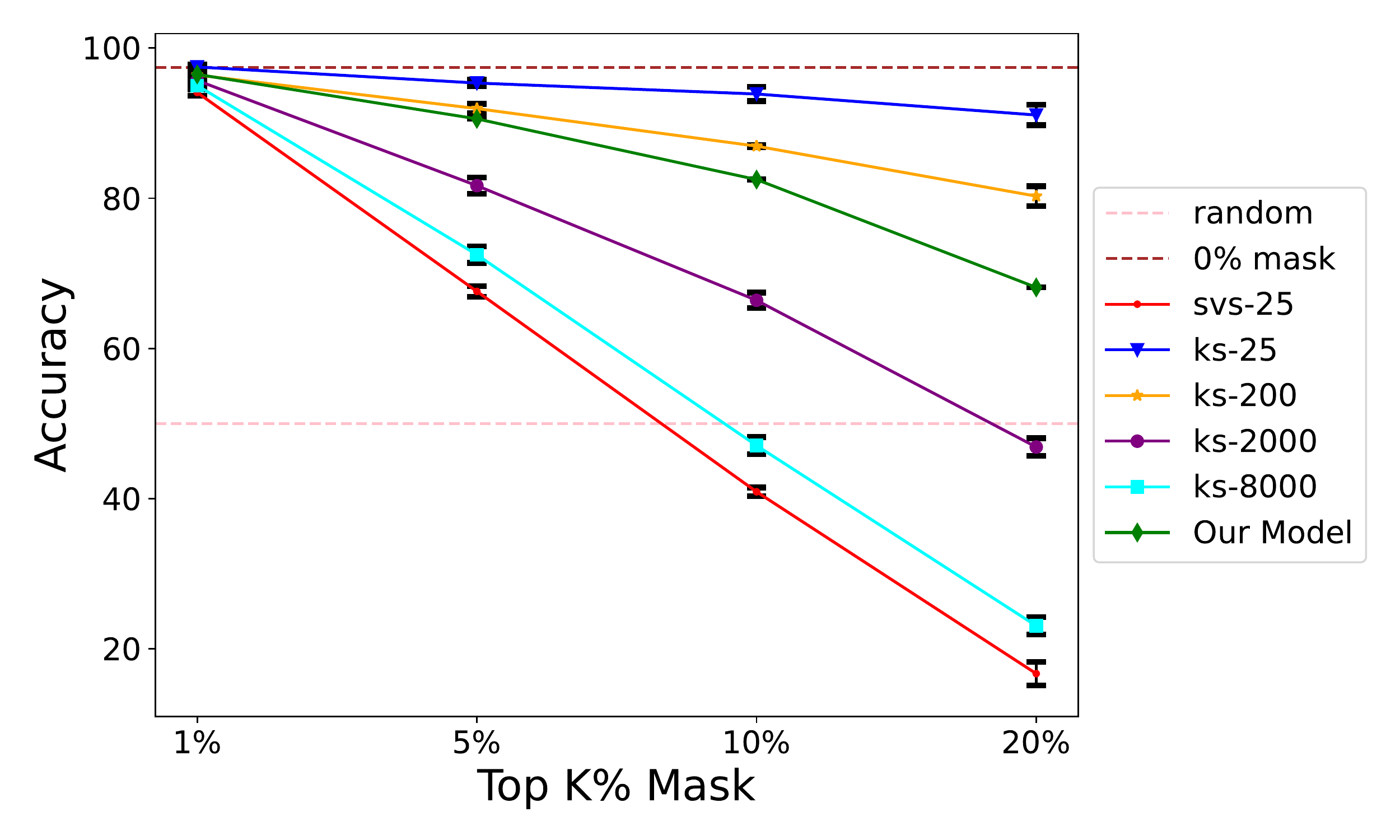}
    \vspace{-3mm}
    \caption{Yelp-Polarity}
    \label{fig: feat_select_yelp_w_amortized}
  \end{subfigure}
  \begin{subfigure}[b]{0.48\textwidth}
  \centering
    \includegraphics[width=0.9\textwidth]{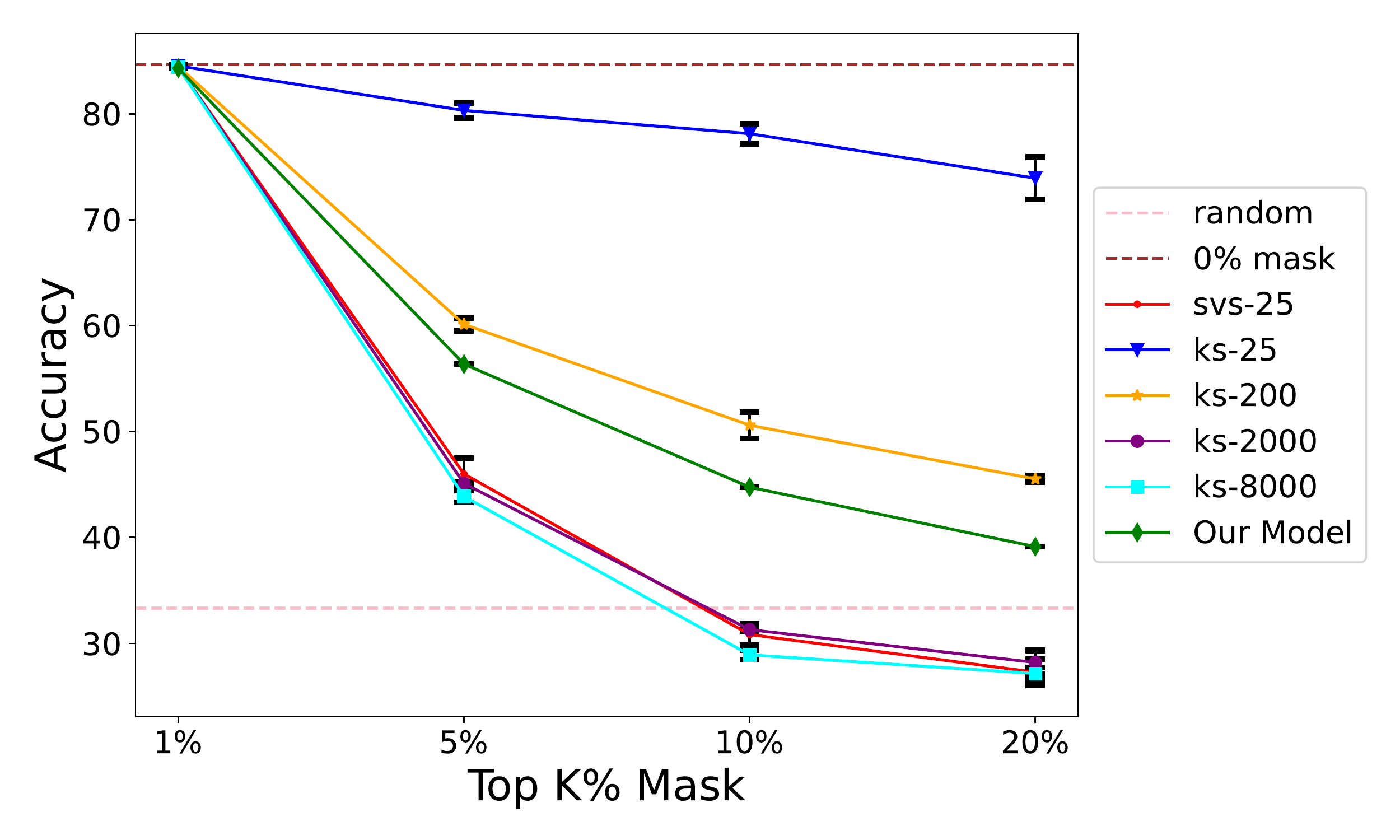}
    \vspace{-3mm}
    \caption{MNLI}
    \label{fig: feat_select_mnli_w_amortized}
  \end{subfigure}
  \caption{Feature selection based on interpretations on Yelp-Polarity and MNLI datasets. The faster the curve drops, the more faithful the explanation scores are. We can see our amortized model is more faithful to the target model compared to KS-200, but are not as faithful as other more costly methods.}
  \label{fig: feat_select_am}
\end{figure*}

\subsection{Evaluating the Quality of Explanation}
\label{sec: downstream_app}
\shortparagraph{Feature Selection.} The first case study is feature selection, which is a straightforward application of local explanation scores. The goal is to find decision-critical features via removing input features gradually according to the  rank given by the explanation methods. Following previous work~\cite{zaidan2007using,jain2019attention,deyoung2020eraser}, we measure faithfulness by changes in the model output after masking tokens identified as important by the explanation method. The more faithful the explanation method is to the target model, the more performance drop will be incurred by masking important tokens.  

We gradually mask Top-$\alpha$ tokens ($\alpha = 1\%, 5\%, 10\% ,20\%$) and compute the accuracy over corrupted results using the stability evaluation sets for
MNLI and Yelp-Polarity datasets as mentioned in \cref{sec: stability}. 
As the results
show in \cref{fig: feat_select_am},
the amortized model is more faithful
than KS-200
but underperforms KS-2000/8000 and SVS-25. However, the amortized model is more efficient than these methods. So amortized model achieves a better efficiency-faithfulness trade-off.  

\shortparagraph{Explanation for Model Calibration.}  Recent work suggests that good explanations should be informative enough to help users to predict model behavior \citep{doshi2017towards, chandrasekaran2018explanations, hase2020evaluating, ye2021connecting}. \citet{ye2022can} propose to combine the local explanation with pre-defined feature templates (e.g., aggregating explanation scores for overlapping words / POS Tags in NLI as features) 
to calibrate an existing model to new domains. The rationale behind this is that, 
{if the local explanation truly connects to human-understandable model behavior, then following the same way how humans transfer knowledge to new domains, the explanations guided by human heuristics (in the form of feature templates) should help calibrate the model to new domains.}
Inspired by this, we conduct a study using the same calibrator architecture but plugging in different local explanation scores. 

We follow \citet{ye2022can} to calibrate a fine-tuned MNLI model\footnote{\url{https://huggingface.co/textattack/bert-base-uncased-MNLI}} to MRPC. The experiment results are shown in \cref{tab:calib-amortized}.
In the table, ``BOW'' means the baseline that uses constant explanation scores when building the features for the calibration model. Compared with the explanation provided by  KS-2000, the explanation given by the amortized model achieves better accuracy, suggesting that the amortized model learns robust explanation scores that can be generalized to out-of-domain data in downstream applications.\footnote{See \cref{sec: domain-transfer} for a domain transfer experiment that directly compares to SVS-25 and w/o calibration.} 

\begin{table}[htbp!]
\small
\centering
\begin{tabular}{@{}ccH@{}}
\toprule
Model                     & Acc  & AUC  \\ \midrule
BOW                       & 67.3 & 74.6 \\
ShapCal (KS-2000) & 67.4 & 74.3 \\
ShapCal (Amortized)       & 68.0 & 74.5 \\ \bottomrule
\end{tabular}%
\caption{Calibration Experiments for Amortized Models. 
The explanation scores can help the calibrator achieves better accuracy on out-of-domain data
than KS-2000. }
\label{tab:calib-amortized}
\end{table}

\section{Conclusion}
In this paper, we empirically demonstrated that it is challenging to obtain stable explanation scores on long text inputs.
Inspired by the fact that different instances can share similarly important features, we proposed to efficiently estimate the explanation scores through an amortized model trained to fit pre-computed reference explanation scores.

In the future, we plan to explore model architecture and training loss for developing  effective amortized models. In particular, we may incorporate sorting-based loss to learn the ranking order of features. Additionally, we could investigate the transferability of the amortized model across different domains, as well as exploring other SHAP-based methods instead of the time-consuming SVS-25 in the data collection process to improve efficiency further.

\section*{Limitations}
In this paper, we mainly focus on developing an amortized model to efficiently achieve a reliable  estimation of SV. Though not experimented with in the paper, our method can be widely applied to other black-box post-hoc explanation methods including LIME~\cite{ribeiro2016lime}. Also, due to the limited budget, we only run experiments on BERT-based models. However, as we do not make any assumption for the model as other black-box explanation methods, our amortized model can be easily applied to other large language models. We only need to collect the model output and our model can be trained offline with just thousands of examples as we show in our method and experiments. 

\shortparagraph{Comparison and Training with Exact Shapley Values}
Computing exact SV is computationally prohibitive for large language models (LLMs) on lengthy text inputs, as it necessitates the evaluation of LLMs on an exponential (in sequence length) number of perturbation samples per instance. As a result, we resort to using SVS-25, which serves as a reliable approximation, for training our amortized models.

\section*{Acknowledgements}

We want to thank Xi Ye and Prof. Greg Durrett for their help regarding their previous work and implementation on using SV for calibration (\cref{sec: downstream_app}). We thank the generous support from AWS AI on computational resources and external collaborations. We further thank Prof. Chenhao Tan for the high-level idea discussion on explainability stability issues at an early stage of this paper, and thank Prof. Yongchan Kwon and Prof. James Zou for their in-depth theoretical analysis of suboptimality of uniform sampling of computing SV. We thank all anonymous reviewers and chairs at ACL'23 and ICLR'23 for their insightful and helpful comments. Yin and Chang are supported in part by a CISCO grant and a Sloan Fellowship.
HH is supported in part by a Cisco grant and Samsung Research (under the project Next Generation Deep Learning: From Pattern Recognition to AI).
\bibliography{acl2023}
\bibliographystyle{acl_natbib}
\newpage
\appendix

\newpage
\appendix
\section{Adaption for FastSHAP Baseline}
\label{app: fastshap}
As we mentioned in \cref{sec: experiment}, we build our amortized models upon a pre-trained encoder BERT~\citep{devlin2019bert}.
However, using the pre-trained encoder significantly increases the memory footprint when running FastSHAP. 
In particular, we have to host two language models on GPUs, one for the amortized model and the other one for the target model.  
Therefore, we can only adopt the batch size equal to 1 and $32$ perturbation samples per instance. Following the proof in FastSHAP, this is equivalent to teaching the amortized model to approximate KS-32, which is an unreliable interpretation method (See \cref{sec: training_sensitivity}). 

In experiments, we find that the optimization of FastSHAP is unstable. After an extensive hyper-parameter search, we set the learning rate to 1e-6 and increased the number of epochs to $30$. However, this requires us to train the model on a single A100 GPU for 3 days to wait for FastSHAP to converge.

\section{Scientific Artifacts License}
For the datasets used in this paper, MNLI~\cite{williams2018broad} is released under ONAC's license. Yelp-Polarity~\cite{zhangCharacterlevelConvolutionalNetworks2015} and SST-2~\cite{socher-etal-2013-recursive} datasets does not provide detailed licenses. 

For model checkpoints used in this paper, they all come from textattack project~\cite{morris2020textattack} and they are open-sourced under MIT license. 

For implementation, we mainly use Captum~\cite{kokhlikyan2020captum} and Thermostat~\cite{feldhus2021thermostat}. Captum is open-sourced under BSD 3-Clause "New" or "Revised" License and Thermostat is open-sourced under Apache License 2.0.

\section{Training Details}
\label{app: training_details}
In this section, we introduce our dataset preprocessing, hyperparameter settings and how we train the models. 

For both MNLI and Yelp-Polarity datasets, we split them
into 8:1:1 for training, validation, and test sets.

The hyperparameters of amortized models are tuned on the validation set. We use Adam~\citep{kingma2015adam} optimizer with a learning rate of 5e-5, train the model for at most 10 epochs and do early stopping to select best model checkpoints.

\end{document}